\documentclass{article} 
\usepackage{arxiv_version,times}
\usepackage{booktabs}
\usepackage{multirow}
\usepackage[table]{xcolor}
\usepackage{graphicx}
\definecolor{lightgreen}{rgb}{0.8,1,0.8}
\definecolor{lightblue}{rgb}{0.8,0.9,1}

\usepackage{amsmath,amsfonts,bm}

\def\eqref#1{equation~\ref{#1}}

\def\1{\bm{1}}

\DeclareMathAlphabet{\mathsfit}{\encodingdefault}{\sfdefault}{m}{sl}
\SetMathAlphabet{\mathsfit}{bold}{\encodingdefault}{\sfdefault}{bx}{n}

\usepackage{hyperref}
\usepackage{url}
\usepackage{caption}
\usepackage{enumitem}
\usepackage{amsmath}
\usepackage{tabularx}

\title{FILIGREE3D: Scaling Sparse Latent Flow Matching for Ultra-High-Resolution Image-to-3D Generation}

\author{Hongjie Li$^{1*}$, Xinran Yang$^{1*}$, Xiuchao Wu$^{1}$, Jiangjing Lyu$^{1,\dag}$, Chengfei Lv$^{1,\dag}$\\
$^{1}$AlibabaGroup\\
}

\iclrfinalcopy 
\begin{document}

\maketitle

\begin{center}
    \vspace{-5mm}
    \centering
    \includegraphics[width=\linewidth,trim=0 0 0 0,clip]{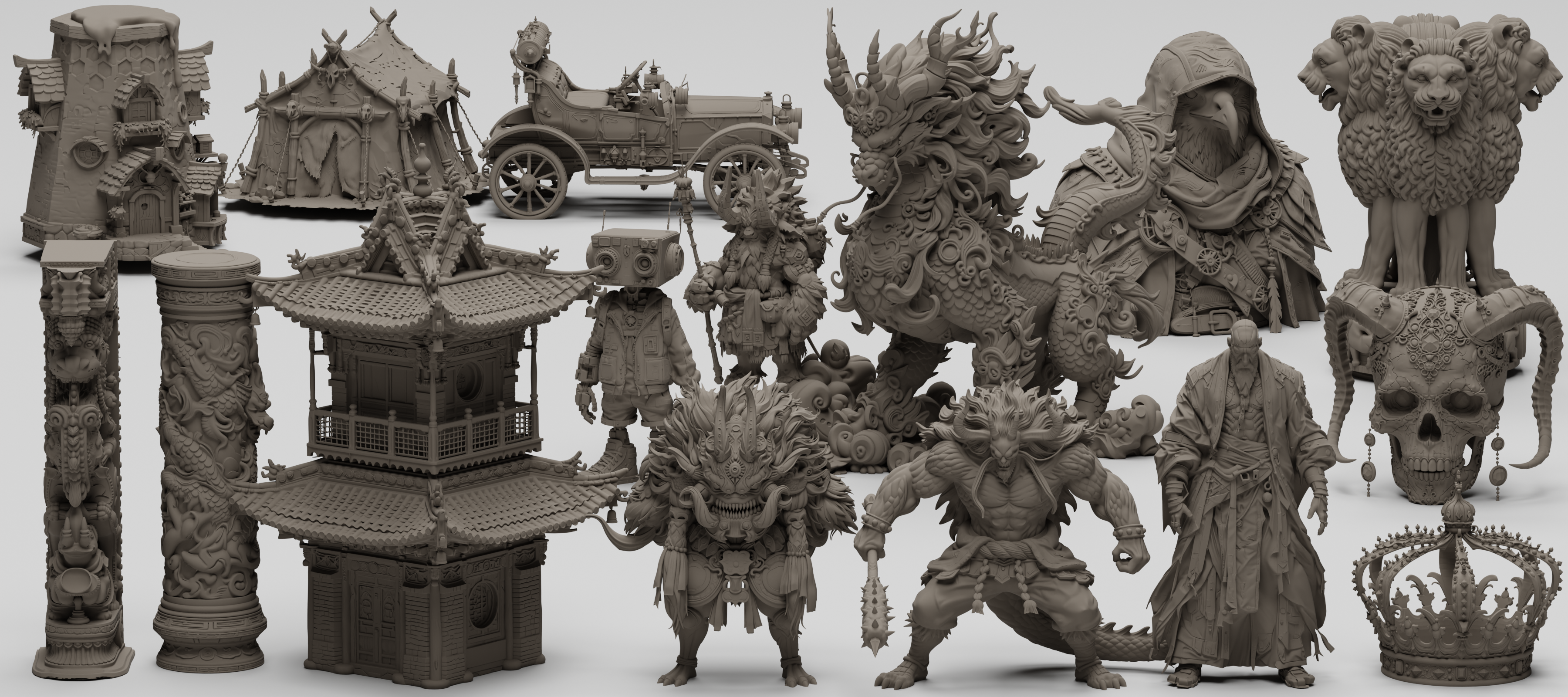}
    \captionof{figure}{\textbf{High-resolution geometry generated by Filigree3D.} Texture-free renders at $2048^3$ highlight thin structures, intricate ornamentation, and sharp surface details.}
    \label{fig:teaser}
\end{center}

\renewcommand{\thefootnote}{} 
\footnotetext{$^*$Equal contribution. $\dag$Corresponding authors.}

\begin{abstract}
Scaling image-to-3D generation to ultra-high resolutions requires controlling rapidly growing computational costs without sacrificing fine geometric detail. We present \textbf{Filigree3D}, a sparse latent flow-matching framework that generates 3D geometry from a single image at voxel resolutions up to $2048^3$, with straightforward extensibility to $4096^3$. To make training tractable, we introduce Structure-Aware Sparse Scaling, which combines spatial bounding with alternating local-global attention to constrain token growth while preserving both fine-scale details and long-range structural context. To enhance detail reconstruction, we curate training samples based on their high-resolution geometric gains and inject multi-scale image features into a sparse 3D DiT, effectively coupling structural semantics with fine-grained visual cues. Furthermore, a visibility-aware voxel regularization strategy improves robustness against sparse perturbations and facilitates the completion of unobserved geometry. Under our default configuration, Filigree3D maintains peak GPU memory consumption within practical limits for contemporary hardware, enabling the generation of highly intricate 3D geometry in approximately one minute. Extensive experiments demonstrate that our method yields substantial improvements in overall geometric fidelity and fine-detail preservation compared to existing baselines, validating practical, detail-preserving 3D generation at unprecedented resolutions.
\end{abstract}

\section{Introduction}
\label{sec:introduction}

Generating a high-fidelity 3D asset from a single image is a longstanding goal in computer vision and graphics, with broad applications in content creation, digital commerce, simulation, and embodied intelligence. Recent image-to-3D systems have made substantial progress by combining pretrained visual encoders, structured 3D representations, and powerful generative priors~\citep{liu2023one,long2024wonder3d,xu2024instantmesh,xiang2025trellis}. Despite these advances, existing models may still oversmooth or omit geometric details, including thin structures, sharp creases, shallow reliefs, and subtle surface variations. A finer 3D representation can raise the upper bound on recoverable detail, but high nominal resolution alone does not guarantee high-fidelity geometry.

Scaling single-image 3D generation to ultra-high resolution exposes three intertwined challenges. First, the computational cost grows rapidly with spatial resolution. Dense voxel grids incur cubic storage and computation, motivating octree and sparse-convolutional representations that avoid allocating features in empty space~\citep{riegler2017octnet,graham20183d,choy20194d,ren2024xcube}. However, even for a sparse surface, the number of active elements grows sharply as the voxel size decreases, making quadratic global self-attention difficult to scale under practical memory budgets. Thus, high-resolution generation requires more than sparsity alone: it must bound fine-scale computation without discarding the local neighborhoods that define geometric detail.

Second, computational locality must not come at the expense of global shape coherence. Spatial cropping and strictly local windowed attention are natural ways to control memory and attention cost, but they limit direct long-range interaction. Shifted-window and focal mechanisms partially compensate for this limitation through cross-window or pooled global context~\citep{liu2021swin,yang2021focal}. Independent crops may additionally introduce artificial boundaries and a discrepancy between partial-support training and complete-object inference. These effects are particularly problematic at high resolution, where local structures must remain consistent with the object-level arrangement. A scalable model must therefore preserve fine local interactions while maintaining communication across the full shape.

Third, high-resolution learning requires informative conditioning and genuinely detail-bearing supervision. Single-view reconstruction is intrinsically ambiguous: visible surfaces are constrained by image evidence, whereas self-occluded regions must be inferred from learned shape priors~\citep{tatarchenko2019single,wu2018learning}. Features from different ViT depths capture complementary positional and semantic information~\citep{amir2021deep}, while DINOv2 provides strong general-purpose visual representations~\citep{oquab2023dinov2}, motivating the use of both intermediate- and final-layer features for conditioning. Training data pose a related challenge: large-scale 3D collections contain diverse objects of uneven quality~\citep{deitke2023objaverse,deitke2023objaversexl}, but scale alone does not guarantee fine-detail supervision. High-resolution training should therefore prioritize assets that reveal additional geometric structure at the target resolution. Thus, both image conditioning and data selection must align with the goal of learning geometric detail.

To address these challenges, we introduce \textbf{Filigree3D}, a sparse latent flow-matching framework for single-image 3D generation at an effective voxel resolution of $2048^3$. Rather than treating resolution as an isolated representational choice, Filigree3D jointly addresses the computational, architectural, conditioning, and data challenges required to translate higher resolution into recoverable geometric detail.  Specifically, a \emph{structure-aware sparse scaling strategy} is proposed to make training tractable while preserving sufficient context and supports full-support inference without crop stitching. Within the sparse 3D DiT, efficient global communication complements fine-grained local interactions, while block-specific multi-level conditioning adapts visual evidence to representations at different network depths. In parallel, resolution-gain data curation prioritizes assets that provide meaningful supervision for fine geometry. Finally, visibility-aware voxel regularization improves robustness to sparse perturbations and facilitates the completion of unobserved regions. Experiments demonstrate that these components effectively translate higher representational resolution into finer recoverable geometry with only modest computational overhead, allowing Filigree3D to outperform existing image-to-3D methods in geometric detail while remaining competitive with commercial systems. Our contributions are summarized as follows:
\begin{itemize}[
    leftmargin=1.5em,
    itemsep=3pt,
    parsep=0pt,
    topsep=2pt,
    partopsep=0pt
]
    \item We present Filigree3D, a sparse latent flow-matching framework operating at an effective voxel resolution of $2048^3$, scalable to $4096^3$, with joint full-support inference.
    \item Structure-Aware Sparse Scaling enables tractable high-resolution training while preserving fine-scale geometric interactions and object-level communication.
    \item Block-specific multi-level feature injection and resolution-gain data curation align visual conditioning and training supervision with fine-detail reconstruction.
\end{itemize}

\section{RELATED WORK}
\label{sec:related_work}

\paragraph{Single-image 3D reconstruction and generation.}
Score-distillation methods optimize a per-instance 3D representation using 2D diffusion priors~\citep{poole2023dreamfusion,liu2023zero123}, achieving high visual quality but requiring minutes to hours per object. Large reconstruction models established a feed-forward alternative~\citep{hong2024lrm,tochilkin2024triposr,xu2024instantmesh}, although their geometric detail remains coupled to the resolution of the tri-plane, implicit field, or mesh-extraction grid. More recent systems learn generative priors over native 3D representations, increasingly under a flow-matching objective that regresses a vector field along a prescribed probability path~\citep{lipman2023flowmatching,liu2023rectifiedflow}. CLAY~\citep{zhang2024clay}, Michelangelo~\citep{zhao2023michelangelo}, TripoSG~\citep{li2025triposg}, and CraftsMan3D~\citep{li2025craftsman3d} scale latent-set or rectified-flow generation to large 3D asset collections; Hunyuan3D~2.1~\citep{hunyuan3d2025} and Step1X-3D~\citep{li2025step1x3d} pair shape VAEs with flow-based DiTs supervised on TSDF grids; Hi3DGen~\citep{ye2025hi3dgen} bridges normal-map estimation and 3D generation; and Direct3D~\citep{wu2024direct3d} introduces a scalable image-to-3D latent diffusion transformer. Pixal3D takes a complementary route: it back-projects multi-scale image features into a pixel-aligned 3D feature volume in the camera frame, conditioning a Direct3D-S2-based sparse SDF backbone trained progressively up to $1024^3$~\citep{li2026pixal3d}. The same flow-matching objective has been paired with non-voxel latent domains, including point-structured Gaussians~\citep{lan2025gaussiananything}, continuous shape tokens~\citep{chang2024shapetokenization}, and topology-bearing sparse voxels~\citep{zhao2026lato}. Beyond research prototypes, these advances have been productized in commercial image-to-3D platforms such as Hunyuan3D~3.1~\citep{hunyuan3d31}, Rodin~v2.5~\citep{rodinv25}, and Tripo~v3.1~\citep{tripov31}, which emphasize production-ready asset and material quality. Filigree3D inherits sparse flow matching for image-conditioned generation but diverges from representation-level modifications. 

\paragraph{Sparse and structured 3D latent representations.}
The choice of 3D latent space largely determines the attainable geometric resolution. Set-based representations compress a surface into a fixed collection of latent vectors and generate them with latent diffusion~\citep{zhang2023shape2vecset,vahdat2022lion,lan2024ln3diff}; recent benchmarking reveals that reconstruction quality and generation diversity remain in tension~\citep{chen2025dora}. These compact tokens do not retain explicit spatial support, limiting resolution scaling. Voxel-based approaches exploit spatial sparsity through hierarchies~\citep{ren2024xcube}, octrees~\citep{xiong2025octfusion}, primitives~\citep{chen20243dtopiaxl}, constructive geometry~\citep{li2025sparc3d}, or part-level attention~\citep{chen2025ultra3d}. TRELLIS introduces structured latents (SLATs) that pair active voxel coordinates with local features, factorizing generation into sparse-structure and latent-feature stages~\citep{xiang2025trellis}, although its mesh decoder operates on a $256^3$ grid. SparseFlex scales its VAE to $1024^3$ with rectified flow~\citep{he2025sparseflex}, while Direct3D-S2 combines a sparse SDF VAE with spatial sparse attention at the same scale~\citep{wu2025direct3ds2}. TRELLIS.2 further adopts the O-Voxel representation and decoupled Sparse Compression VAEs, trained up to $1024^3$ with $1536^3$ assets via test-time cascaded scaling~\citep{xiang2026native}. LATTICE instead anchors a compact VoxSet on a coarse sparse grid with a coordinate-aware rectified-flow transformer~\citep{lai2026lattice}. Together, these methods highlight the scalability benefits of spatial sparsity, while also exposing the distinction among latent density, decoder resolution, and nominal voxel resolution. Filigree3D targets effective resolutions beyond $1024^3$, addressing the scaling of the sparse generative backbone itself.

\section{METHOD}

\label{sec:preliminaries}
\begin{figure}[t]
    \centering
    \includegraphics[width=\linewidth]{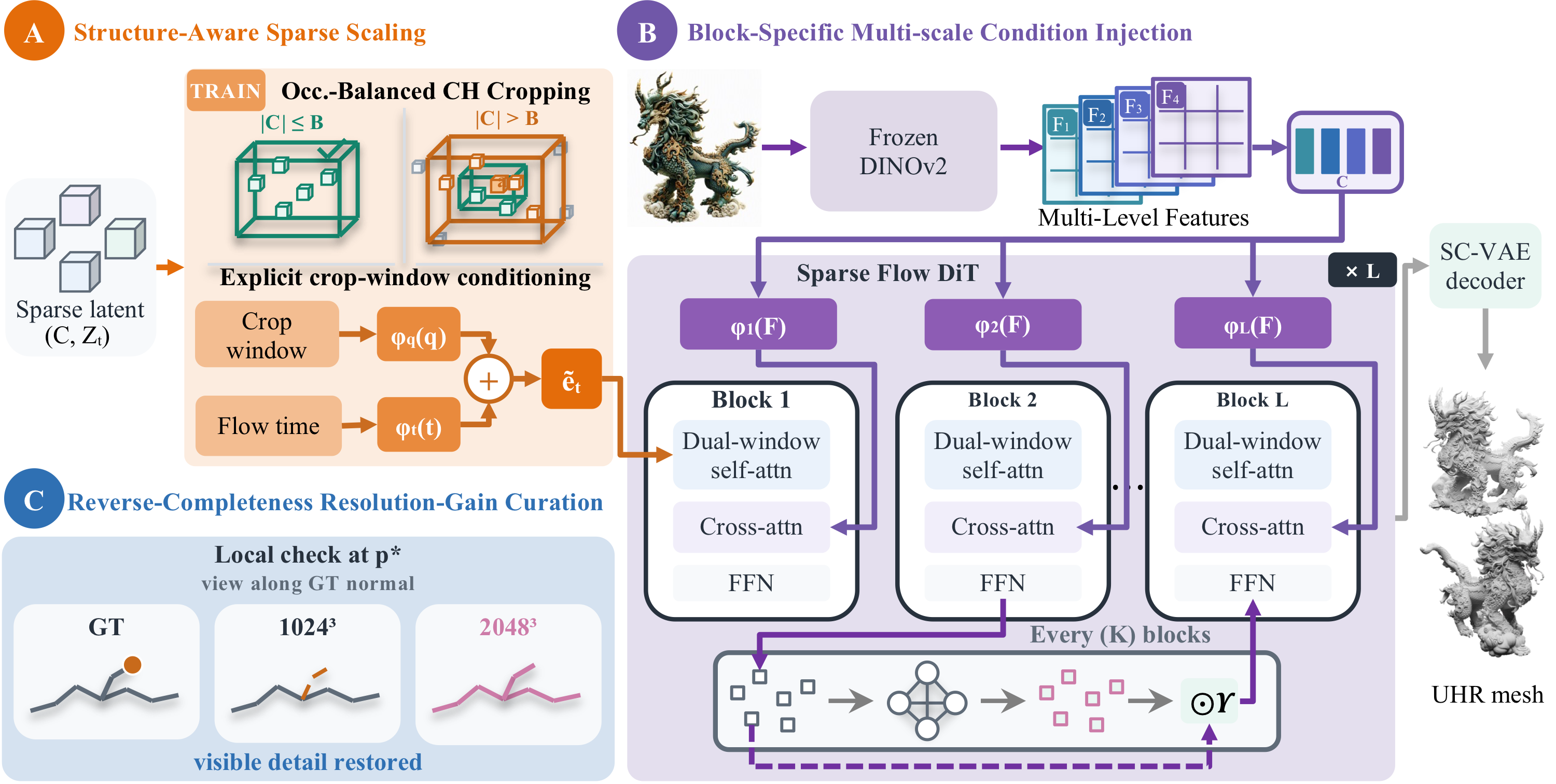}
    \caption{\textbf{Overview of Filigree3D.} The framework combines (A) structure-aware sparse scaling for budgeted processing, (B) block-specific multi-scale condition injection for hierarchical image guidance, and (C) resolution-gain curation for resolution-sensitive samples.}
    \label{fig:overview of Filigree3D}
\end{figure}

\subsection{PRELIMINARY}
Filigree3D builds on the O-Voxel representation and sparse generative framework of TRELLIS.2~\citep{xiang2026native}. We briefly review the relevant formulations below.

\paragraph{O-Voxel and Sparse flow matching.}
At resolution $N^3$, O-Voxel represents geometry using only $L$ surface-intersecting voxels, denoted by $\mathcal{O}=\{(\mathbf{p}_i,\mathbf{f}_i)\}_{i=1}^{L}$. Each active coordinate $\mathbf{p}_i\in\{0,\dots,N-1\}^{3}$ is paired with $\mathbf{f}_i=(\mathbf{v}_i,\boldsymbol{\delta}_i,\gamma_i)$, which encodes the dual vertex, surface intersections, and triangulation parameter. Flexible Dual Grid converts $\mathcal{O}$ into a triangle mesh. For efficient latent modeling, TRELLIS.2 employs an SC-VAE to encode it as $\mathcal{Z}=E(\mathcal{O})$. The resulting latent comprises $M$ active entries $(\mathbf{q}_j,\mathbf{z}_j)$, where $\mathbf{q}_j\in\{0,\dots,N/16-1\}^{3}$ and $\mathbf{z}_j\in\mathbb{R}^{32}$. The encoder reduces each spatial dimension by a factor of $16$, while the decoder reconstructs $\mathcal{O}$ from $\mathcal{Z}$ for subsequent mesh extraction.
Let $\mathcal{Q}=\{\mathbf{q}_j\}_{j=1}^{M}$ be the active support predicted by the sparse-structure stage and $\mathbf{x}_0=[\mathbf{z}_1,\ldots,\mathbf{z}_M]$ the corresponding latent features. With $\mathcal{Q}$ fixed, an image-conditioned sparse DiT learns to transport noisy features toward $\mathbf{x}_0$:
\begin{equation}
\begin{aligned}
    \mathbf{x}_t
    &= (1-t)\mathbf{x}_0
    + \left[\sigma_{\min}+(1-\sigma_{\min})t\right]\boldsymbol{\epsilon}, \\
    \mathcal{L}_{\mathrm{CFM}}
    &= \mathbb{E}
      \left[
        \left\|
          \mathbf{v}_{\theta}(\mathcal{Q},\mathbf{x}_t,t,\mathbf{y})
          - \left((1-\sigma_{\min})\boldsymbol{\epsilon}-\mathbf{x}_0\right)
        \right\|_2^2
      \right],
\end{aligned}
\label{eq:trellis_cfm}
\end{equation}
where $\boldsymbol{\epsilon}\sim\mathcal{N}(\mathbf{0},\mathbf{I})$, $t\sim\mathcal{U}(0,1)$, $\mathbf{y}$ denotes the image condition, and $\sigma_{\min}=10^{-5}$. At inference, the learned flow is integrated from noise to the data endpoint, and the resulting latent is decoded into a mesh. The following sections describe our extensions for ultra-high-resolution generation.

\paragraph{Overview.} Given a single image, Filigree3D generates high-resolution 3D geometry via sparse latent flow matching. As shown in Figure~\ref{fig:overview of Filigree3D}, an image-conditioned sparse denoiser predicts latents that are decoded by a resolution-specific SC-VAE at an effective $2048^3$ resolution, scalable to $4096^3$. Structure-aware sparse processing bounds token cost while preserving local detail and global context, while block-specific conditioning provides stage-adaptive guidance. Resolution-gain curation prepares training data by retaining assets that reveal additional geometry at higher resolutions. 

\subsection{Structure-Aware Sparse Scaling}
\label{sec:structure_aware_sparse_scaling}

Although O-Voxel latents are sparse, the number of occupied tokens still grows substantially with output resolution and may exceed a fixed per-sample budget $B$ under global attention. The central challenge is therefore to train under a bounded token count without sacrificing spatial coverage or object-level communication. We address this challenge with \emph{Structure-Aware Sparse Scaling}, which combines occupancy-balanced core--halo cropping and explicit crop-window conditioning for budgeted training with periodic coarse-global interaction for long-range information exchange. Let $\mathcal{X}_t=(\mathcal{C},\mathbf{Z}_t)$ denote a sparse latent, where $\mathcal{C}\subseteq\{0,\ldots,R-1\}^3$ is the occupied support and $\mathbf{Z}_t$ contains its time-dependent features. If $|\mathcal{C}|\leq B$, we retain the complete support. Otherwise, we select a spatially coherent subset containing at most $B$ tokens before sampling the flow time and noise. Once this subset is selected, its coordinates remain unchanged within the flow-matching objective, while only the associated features are perturbed and predicted.

\paragraph{Occupancy-balanced core--halo cropping.}
Sampling anchors uniformly from occupied tokens biases training toward geometrically dense regions. We instead partition the latent grid into coarse cells of width $g$, uniformly sample a non-empty cell, and then sample an occupied anchor $\mathbf{a}$ within that cell. Around the anchor, we define an outer context region and an inner supervision core:
\begin{equation}
\begin{aligned}
    \mathcal{I}_{\mathrm{ctx}}(s;\mathbf{a})
    &=\left\{i:\lVert\mathbf{c}_i-\mathbf{a}\rVert_{\infty}\leq s\right\},\\
    \mathcal{I}_{\mathrm{core}}(s;\mathbf{a})
    &=\left\{i:\lVert\mathbf{c}_i-\mathbf{a}\rVert_{\infty}
    \leq \max(s-h,0)\right\},
\end{aligned}
\label{eq:core_halo_regions}
\end{equation}
where $h$ is the halo width. The context radius adapts to local occupancy within a fixed token budget:
\begin{equation}
    s^{\star}
    =\max\left\{s\in\mathbb{Z}_{\geq 0}:
    |\mathcal{I}_{\mathrm{ctx}}(s;\mathbf{a})|\leq B\right\}.
    \label{eq:occupancy_adaptive_crop}
\end{equation}
The denoiser processes the full context, while the flow-matching loss is restricted to the core:
\begin{equation}
    \mathcal{L}_{\mathrm{crop}}
    =
    \left|\mathcal{I}_{\mathrm{core}}(s^\star;\mathbf{a})\right|^{-1}
    \sum\nolimits_{i\in\mathcal{I}_{\mathrm{core}}(s^\star;\mathbf{a})}
    \lVert\widehat{\mathbf{v}}_i-\mathbf{v}_i\rVert_2^2.
    \label{eq:core_only_supervision}
\end{equation}
This separation preserves contextual evidence while avoiding direct supervision near artificial crop boundaries. Unlike a fixed spatial crop, the selected extent expands in sparse regions and contracts in dense regions to use the available budget. Coordinates remain in the original latent-grid frame, and samples satisfying $|\mathcal{C}|\leq B$ use complete-support supervision.

\paragraph{Explicit crop-window conditioning.}
A local crop and a complete object may contain identical absolute coordinates but provide different spatial evidence. Conditioning only on occupied-token extrema cannot resolve this ambiguity because those extrema describe the observed geometry rather than the window from which it was sampled. We therefore encode the actual crop bounds. For a cropped support,
$\boldsymbol{\ell}=\max(\mathbf{0},\mathbf{a}-s^\star\mathbf{1})$ and
$\mathbf{u}=\min((R-1)\mathbf{1},\mathbf{a}+s^\star\mathbf{1})$;
for complete support, $\boldsymbol{\ell}=\mathbf{0}$ and $\mathbf{u}=(R-1)\mathbf{1}$. The normalized window center, size, and crop indicator are
\begin{equation}
    \mathbf{q}
    =
    \left[
        [2(R-1)]^{-1}(\boldsymbol{\ell}+\mathbf{u})\,;
        R^{-1}(\mathbf{u}-\boldsymbol{\ell}+\mathbf{1})\,;
        \rho
    \right],
    \qquad \rho\in\{0,1\}.
    \label{eq:crop_window_descriptor}
\end{equation}
where $\rho=1$ denotes a crop and $\rho=0$ complete support. A two-layer MLP $\phi_q$ injects this descriptor through the flow-time embedding:
\begin{equation}
    \widetilde{\mathbf{e}}_t
    =\phi_t(t)+\phi_q(\mathbf{q}).
    \label{eq:crop_window_conditioning}
\end{equation}
The final layer of $\phi_q$ is zero-initialized so that the added condition does not perturb the original pathway at initialization. This explicit descriptor aligns budgeted crop training with complete-support inference by informing the denoiser of the spatial extent actually observed.

\paragraph{Local interaction with periodic coarse-global communication.}
Windowed sparse attention provides bounded fine-scale computation but restricts communication between distant occupied regions. We use complementary regular and half-window-shifted partitions for local mixing: half of the heads attend within regular windows of width $w$, and the remaining heads use windows shifted by $w/2$ along each axis. To restore object-level context, every $K$ blocks we group fine tokens by stride-$p$ coarse coordinates,
\begin{equation}
    \mathbf{u}_i
    =\left\lfloor p^{-1}\mathbf{c}_i\right\rfloor,
    \qquad
    \mathcal{S}_{\mathbf{u}}
    =\{i:\mathbf{u}_i=\mathbf{u}\}.
\end{equation}
and aggregate each occupied coarse cell:
\begin{equation}
    \mathbf{g}_{\mathbf{u}}
    =
    \left(1/|\mathcal{S}_{\mathbf{u}}|\right)
    \sum\nolimits_{i\in\mathcal{S}_{\mathbf{u}}}\mathbf{h}_i.
    \label{eq:coarse_pooling}
\end{equation}
Full self-attention is applied only to the resulting $M$ occupied coarse tokens. The coarse context is mapped back via the inverse grouping assignment and fused with the corresponding fine tokens:
\begin{equation}
\widehat{\mathbf{G}}=\operatorname{Attn}_{\mathrm{global}}\!\left(\operatorname{LN}(\mathbf{G})\right),\qquad
\mathbf{h}'_i=\mathbf{h}_i+\boldsymbol{\gamma}\odot\widehat{\mathbf{g}}_{\mathbf{u}_i},\qquad
\boldsymbol{\gamma}\big|_{\mathrm{init}}=\mathbf{0}.
\label{eq:coarse_global_fusion}
\end{equation}
This periodic pathway propagates object-level context without quadratic attention over all fine tokens, while a zero-initialized gate gradually introduces coarse context during training.

\subsection{Block-Specific Multi-Level Condition Injection}
\label{sec:multi_scale_feature_injection}

Effective cross-attention requires conditioning features that capture both fine-grained image details and high-level semantics. However, relying solely on the final encoder layer may omit informative cues retained in intermediate representations. We therefore extract normalized features from four levels of a frozen DINOv2 encoder~\citep{oquab2023dinov2}, denoted by $\{E_{\ell_i}(\mathbf{I})\}_{i=1}^{4}$. Since these features share the same token layout, we concatenate them along the channel dimension to preserve their complementary information without imposing fixed weights. As the sparse voxel representation evolves across the DiT blocks, the most useful combination of abstraction levels may vary with network depth. We thus introduce an independent fusion module for each block:
\begin{equation}
\mathbf{F}^{(l)}_{\mathrm{cond}}
=
\phi_l\!\left(
\operatorname{Concat}_{\mathrm{ch}}
\left[\operatorname{LN}\!\left(E_{\ell_k}(\mathbf{I})\right)\right]_{k=1}^{4}
\right)
\in\mathbb{R}^{P\times D},
\qquad l=1,\ldots,L.
\label{eq:multi_level_condition_injection}
\end{equation}
Here, $\phi_l$ is a block-specific fusion module consisting of layer normalization followed by a two-layer MLP with a GELU activation. This design allows each DiT block to adaptively combine visual cues from different abstraction levels while keeping the number of conditioning tokens fixed at $P$.

\subsection{Visibility-Aware Voxel Regularization}
\label{sec:visibility_aware_voxel_regularization}
To regularize ambiguous occluded geometry, we estimate voxel visibility by comparing the projected voxel depth $d_i$ with a low-resolution depth buffer $D$:
\begin{equation}
    m_i =
    \mathbf{1}[z_i<0]\,
    \mathbf{1}[\pi_i\in\Omega]\,
    \mathbf{1}\!\left[|d_i-D(\pi_i)|<2/R\right],
    \label{eq:voxel_visibility}
\end{equation}
where $\pi_i$ is the projected voxel position, $\Omega$ denotes the image crop, and $R$ is the latent resolution. Unseen voxels ($m_i=0$) receive stronger perturbation and feature masking. We define their regression weights as $w_i=1$ for visible voxels and $w_i=\lambda_{\mathrm{u}}$ for unseen voxels, where $\lambda_{\mathrm{u}}>1$. The regularization objective is
\begin{equation}
\mathcal{L}_{\mathrm{reg}}
=
\frac{\sum_{i\in\mathcal{S}} w_i
\lVert\hat{\mathbf{v}}_i-\mathbf{v}_i\rVert_2^2}
{C\sum_{i\in\mathcal{S}}w_i}
+
\frac{\lambda_{\mathrm{lap}}
\sum_{(i,j)\in\mathcal{E}_{\mathrm{u}}}
\lVert\hat{\mathbf{v}}_i-\hat{\mathbf{v}}_j\rVert_2^2}
{C\left(|\mathcal{E}_{\mathrm{u}}|+\epsilon\right)}.
\label{eq:visibility_regularization}
\end{equation}
Here, $\mathcal{S}$ contains occupied voxels, $\mathcal{E}_{\mathrm{u}}$ contains adjacent unseen-voxel pairs, and $C$ is the feature dimension. This regularization is applied only during training and introduces no inference overhead.

\subsection{Resolution-Gain-Aware Data Curation.}
Effective high-resolution geometry generation requires training assets that reveal additional geometric details as the representation resolution increases. We therefore propose a resolution-gain-aware curation protocol to identify such assets, as illustrated in Figure~\ref{fig:overview of Filigree3D}(c). Given a ground-truth (GT) mesh, we construct its O-Voxel representations at low and high resolutions, reconstruct each using the corresponding SC-VAE, and extract the resulting meshes $\mathcal{M}_{R_{\mathrm{lo}}}$ and $\mathcal{M}_{R_{\mathrm{hi}}}$. To capture small yet important structures, we draw (N) GT surface points using mixed area-based and detail-aware sampling~\citep{chen2025dora}:
\begin{equation}
    p(\mathbf{x})
    =
    \lambda p_{\mathrm{area}}(\mathbf{x})
    +
    (1-\lambda)p_{\mathrm{detail}}(\mathbf{x}),
\end{equation}
where $p_{\mathrm{detail}}$ assigns higher probabilities to regions with large curvature or normal variation. The same GT samples are shared across both resolutions for a fair comparison. Let $d_R(p)$ denote the distance from a GT point $p$ to the surface of $\mathcal{M}_R$. We measure the missing geometry at resolution $R$ by $E_R^{\max}=\max_i d_R(p_i)$ and define the resolution gain as $G_{\mathrm{res}}=E_{R_{\mathrm{lo}}}^{\max}/E_{R_{\mathrm{hi}}}^{\max}$. An asset is selected as a candidate if $E_{R_{\mathrm{lo}}}^{\max}>\tau_{\mathrm{miss}}$ and $G_{\mathrm{res}}>\gamma$. The first condition identifies meaningful geometry missing at low resolution, while the second verifies that it is substantially recovered at high resolution. Because the maximum distance can be sensitive to isolated errors, these criteria are used only for candidate screening. We then inspect aligned reconstructions around the worst-covered GT point and retain only assets showing clear and spatially coherent geometric recovery.

\section{EXPERIMENTS}
\subsection{EXPERIMENT SETUP}
\paragraph{Datasets, Baselines, and Metrics.}
(1) For a fair comparison, we evaluate all methods on the same subset of Toys4K~\citep{stojanov2021using}, comprising 537 manually selected objects with rich geometric detail. To complement the limited coverage of highly intricate shapes in Toys4K, we further introduce FineGeo3D, a supplementary evaluation set of 120 detail-rich cases. (2) We compare Filigree3D quantitatively and qualitatively with Direct3D-S2~\citep{wu2024direct3d}, Hunyuan3D 2.1~\citep{hunyuan3d2025}, TRELLIS.2~\citep{xiang2026native}, and Pixal3D~\citep{li2026pixal3d}, using identical input images. Hunyuan3D 2.1 decodes geometry from a resolution-$384$ octree with 50 diffusion steps; Direct3D-S2 reconstructs $1024^3$ signed distance fields through a dense-to-sparse cascade; and TRELLIS.2 and Pixal3D progressively refine sparse structures and shape latents from $512$ to $1536$. We additionally provide qualitative comparisons with state-of-the-art commercial systems, including Hunyuan3D-3.1~\citep{hunyuan3d31,lai2026lattice}, Rodin v2.5~\citep{rodinv25}, and Tripo v3.1~\citep{tripov31}, to assess fine-detail generation in practical settings. (3) We evaluate geometric fidelity using Chamfer Distance (CD) and F-scores at thresholds of $0.01$ and $0.002$, with the latter emphasizing fine geometric details. Image--geometry alignment is measured by ULIP-2~\citep{xue2024ulip} and Uni3D~\citep{zhou2024uni3d} similarities, where higher scores indicate better semantic consistency. All methods use identical normalization and sampling protocols.
\begin{figure}[t]
    \centering
    \includegraphics[width=\linewidth]{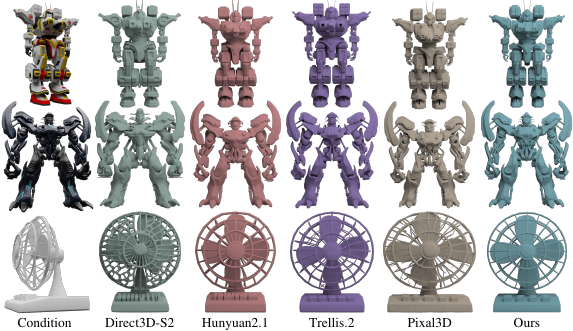}
    \caption{\textbf{Qualitative comparison on Toys4K dataset.}}
    \label{fig:qualitative_Toys4k}
\end{figure}
\begin{table*}[t]
  \centering
  \caption{\textbf{Quantitative comparison on Toys4K and FineGeo3D dataset.} T means inference time.}
  \label{tab:comparisons}

  \small
  \setlength{\tabcolsep}{2.2pt}
  \renewcommand{\arraystretch}{1.05}

  \resizebox{\linewidth}{!}{\begin{tabular}{l c ccccc ccccc}
    \toprule
    \multirow{2}{*}{\textbf{Method}} &
    \multirow{2}{*}{\textbf{T (s)}$\downarrow$} &
    \multicolumn{5}{c}{\textbf{Toys4K}} &
    \multicolumn{5}{c}{\textbf{FineGeo3D}} \\
    \cmidrule(lr){3-7}\cmidrule(lr){8-12}

    & &
    CD$\downarrow$ &
    F1-0.01$\uparrow$ &
    F1-0.002$\uparrow$ &
    ULIP-2$\uparrow$ &
    Uni3D$\uparrow$ &
    CD$\downarrow$ &
    F1-0.01$\uparrow$ &
    F1-0.002$\uparrow$ &
    ULIP-2$\uparrow$ &
    Uni3D$\uparrow$ \\
    \midrule

    Direct3D-S2&69.2 &0.0339 &0.5062 &0.0596 &0.2293 &0.2785 &0.0326 &0.4917 &0.0499 &0.2184 &0.2783 \\
    Hy3D-2.1&17.1 &0.0233&0.6683 & 0.1011&0.2390 & 0.2797&0.0278 &0.5809 &0.0674 &0.2330 &0.2879  \\
    TRELLIS.2 &59.5 &0.0270 & 0.6315&0.0947 &0.2275 &0.2807 &0.0274 &0.5847 &0.0639 &0.2239 &0.2716\\
    Pixal3D &26.2 &0.0360 &0.5542 &0.0888 &0.2080 &0.2775 &0.0345 &0.5448 &0.0674 &0.2115 &0.2717 \\
    \midrule
    \rowcolor{lightgreen}
    Ours-1536 &50.9 &0.0058 &\textbf{0.9744} &0.4616 &0.2624 &0.2965 &0.0054 &0.9963 &0.3679 &0.2665 &0.3042 \\

    \rowcolor{lightblue}
    Ours-2048 &84.2 &\textbf{0.0055} &0.9739 &\textbf{0.5202} &\textbf{0.2724} &\textbf{0.3070} &\textbf{0.0049} &\textbf{0.9969} &\textbf{0.4309} &\textbf{0.2760} &\textbf{0.3126} \\
    \bottomrule
  \end{tabular}
}
\end{table*}

\paragraph{Implementation Details.}
\vspace{-5mm}
(1) \textbf{SC-VAE.}We fine-tune a separate SC-VAE for each target resolution and address resolution-specific issues, including long-edge artifacts at $2048^3$ and coordinate wraparound and out-of-memory failures at $4096^3$. (2) \textbf{DiT.} We set the token budget to (B=50{,}000) and apply global information flow every five Transformer blocks. For multi-level conditioning, we concatenate DINOv2 features from layers 5, 7, 11, and 23 along the channel dimension~\citep{he2025lam}.
(3) \textbf{Data Curation.}
Each GT mesh is normalized to unit maximum extent and reconstructed at $R_{\mathrm{lo}}=1024$ and $R_{\mathrm{hi}}=2048$ using resolution-specific SC-VAEs and Flexible Dual Grid. We set $\lambda=0.7$, use $100{,}000$ shared surface samples for screening and $500{,}000$ for verification, and set $\tau_{\mathrm{miss}}=[2R_{\mathrm{lo}}]^{-1}$, $\gamma=2$, and $\epsilon=0.1R_{\mathrm{hi}}^{-1}$.
(4) \textbf{Training.} To train the model, we use TRELLIS-500K~\cite{xiang2025trellis} with an additional filtering step to retain geometrically complex objects, and we also collect 200K high-quality, high-precision data. Our model is trained for 10 days on 16 NVIDIA H20 GPUs with a batch size of 1. We use AdamW with an initial learning rate of $1\times10^{-4}$.

\subsection{EXPERIMENT RESULTS}
\begin{figure}[t]
    \centering
    \includegraphics[width=\linewidth]{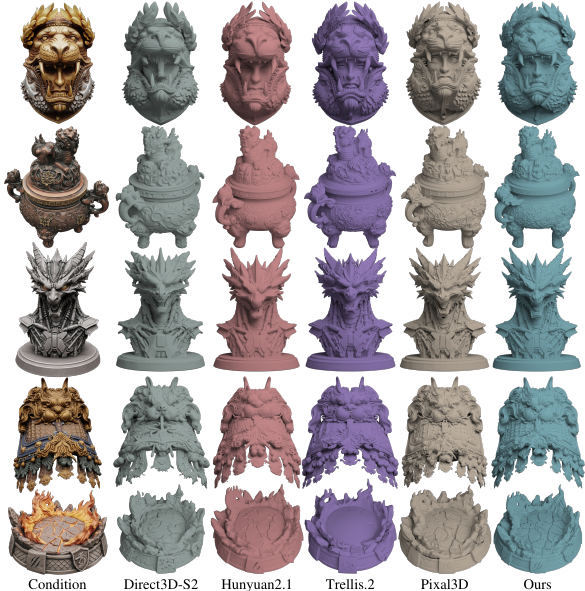}
    \caption{\textbf{Qualitative comparison on FineGeo3D dataset.} }
    \label{fig:qualitative_FineGeo3D}
\end{figure}

\paragraph{Qualitative comparison.} Figures~\ref{fig:qualitative_Toys4k} and~\ref{fig:qualitative_FineGeo3D} compare meshes generated by Filigree3D with those produced by state-of-the-art open-source image-to-3D methods on Toys4K and FineGeo3D, respectively. Filigree3D reconstructs more faithful geometry while preserving finer structures, including thin bars, hollow ornaments, sharp creases, and shallow reliefs. In contrast, existing methods often merge nearby components or oversmooth subtle details, leading to loss of local shape information. Filigree3D also produces more plausible geometry in occluded or partially visible regions, while still retaining rich details in visible regions. These improvements are attributed to our dedicated design for ultra-high-resolution generation and fine-detail preservation.

\paragraph{Quantitative comparison.} Table~\ref{tab:comparisons} reports the quantitative evaluation on the Toys4K subset and FineGeo3D, covering geometric fidelity (CD, F1-0.01, and F1-0.002), semantic alignment (ULIP-2 and Uni3D), and inference time. The runtime reported in the table refers to the inference time of the geometry generation model on a single H20 GPU. Filigree3D consistently outperforms all compared baselines on most metrics across both benchmarks, with particularly large gains on fidelity metrics that emphasize fine surface structures. These results demonstrate that Filigree3D substantially improves geometric reconstruction quality without compromising semantic alignment.





\subsection{ABLATION STUDY}

\begin{table*}[t]
  \centering
  \caption{\textbf{Ablation study on FineGeo3D dataset.}
  GIF, VR, MSF, and DC denote global information flow, voxel regularization, multi-scale features
  and data curation, respectively.}
  \label{tab:ablation}

  \small
  \setlength{\tabcolsep}{6pt}
  \renewcommand{\arraystretch}{1.05}

  \begin{tabular}{lccccc}
    \toprule
    \textbf{Method} &
    CD$\downarrow$ &
    F1-0.01$\uparrow$ &
    F1-0.002$\uparrow$ &
    ULIP-2$\uparrow$ &
    Uni3D$\uparrow$ \\
    \midrule

    w/o GIF  &0.0053 &0.9865 &0.4121 &0.2605 &0.3077 \\
    w/o VR   & 0.0052& 0.9863& 0.4062& 0.2610&0.3092 \\
    w/o MSF   & 0.0072& 0.9662& 0.3845& 0.2407&0.2712 \\
    w/o DC   & 0.0060& 0.9765& 0.3941& 0.2509&0.2915 \\
    
    \midrule

    \rowcolor{lightgreen}
    Full Pipeline &\textbf{0.0049} &\textbf{0.9969} &\textbf{0.4309} &\textbf{0.2760} &\textbf{0.3126} \\

    \bottomrule
  \end{tabular}
\end{table*}

\paragraph{Component Analysis.}
As shown in Table~\ref{tab:ablation}, removing multi-scale features (MSF), global information flow (GIF), voxel regularization (VR), or data curation (DC) consistently degrades performance on FineGeo3D. The full model achieves the best results, demonstrating their complementary contributions to geometric fidelity and semantic consistency.

\paragraph{Resolution Gains and Commercial Comparisons.}
\begin{figure}[t]
    \centering
    \includegraphics[width=\linewidth]{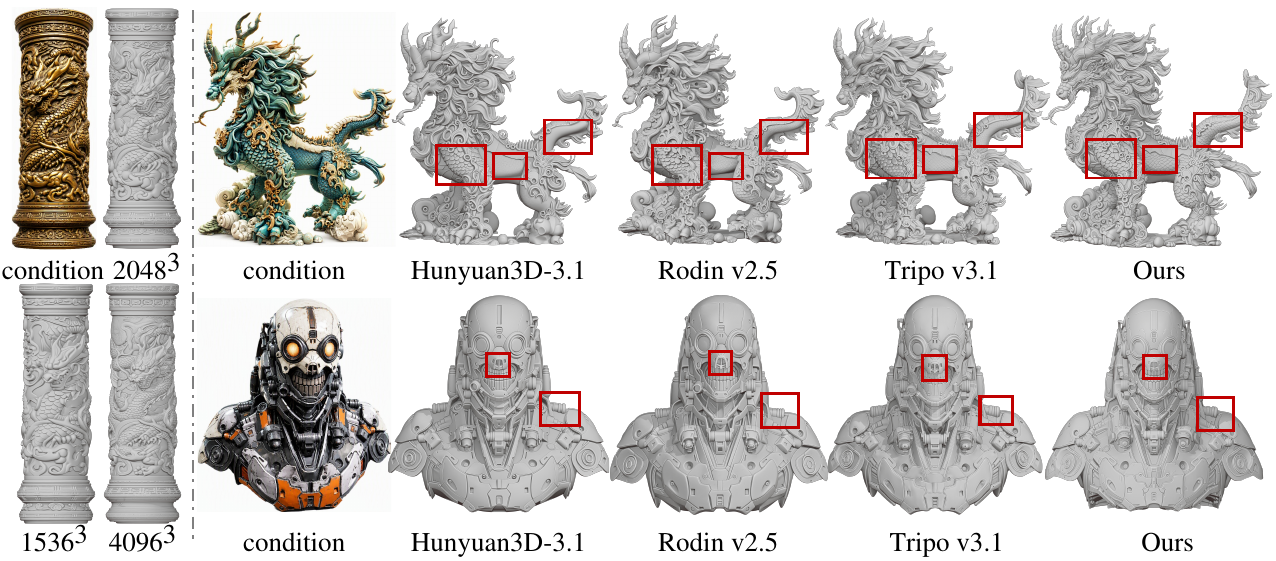}
\caption{\textbf{Resolution Scaling and Commercial Comparison.} Left: Filigree3D recovers increasingly fine geometry from  $1536^3$ to $4096^3$. Right: comparison with leading commercial systems at their highest-quality settings. Red boxes highlight fine geometric details.}
\label{fig:resolution_gain}
\end{figure}
Figure~\ref{fig:resolution_gain} presents qualitative comparisons along two dimensions. First, increasing the generation resolution from $1536^3$ to $2048^3$ and $4096^3$ progressively improves geometric fidelity, yielding sharper relief patterns and more complete fine-scale structures. Second, compared with commercial systems including Hunyuan3D-3.1~\citep{hunyuan3d31,lai2026lattice}, Rodin v2.5~\citep{rodinv25}, and Tripo v3.1 ~\citep{tripov31}, Filigree3D better preserves intricate geometric patterns and high-frequency details, as highlighted by the red boxes. These results demonstrate both the favorable resolution scalability of Filigree3D and its competitive performance in fine-grained geometry generation.

\section{Conclusion}
\label{sec:conclusion}


We presented \textbf{Filigree3D}, a sparse latent flow-matching framework for single-image 3D generation at an effective $2048^3$ resolution, scalable to $4096^3$. By jointly designing sparse computation, multi-level conditioning, and resolution-aware supervision, Filigree3D recovers finer geometry while preserving global consistency under a fixed token budget. Experiments show finer details with modest overhead and performance competitive with leading commercial systems. We hope this work supports further advances in high-resolution 3D generation.

\bibliography{iclr2027_conference}

@inproceedings{poole2023dreamfusion,
  title     = {{DreamFusion}: Text-to-{3D} using {2D} Diffusion},
  author    = {Poole, Ben and Jain, Ajay and Barron, Jonathan T. and Mildenhall, Ben},
  booktitle = {International Conference on Learning Representations},
  year      = {2023}
}

@inproceedings{liu2023zero123,
  title     = {Zero-1-to-3: Zero-shot One Image to {3D} Object},
  author    = {Liu, Ruoshi and Wu, Rundi and Van Hoorick, Basile and Tokmakov, Pavel and Zakharov, Sergey and Vondrick, Carl},
  booktitle = {IEEE/CVF International Conference on Computer Vision},
  year      = {2023}
}

@inproceedings{hong2024lrm,
  title     = {{LRM}: Large Reconstruction Model for Single Image to {3D}},
  author    = {Hong, Yicong and Zhang, Kai and Gu, Jiuxiang and Bi, Sai and Zhou, Yang and Liu, Difan and Liu, Feng and Sunkavalli, Kalyan and Bui, Trung and Tan, Hao},
  booktitle = {International Conference on Learning Representations},
  year      = {2024}
}

@article{tochilkin2024triposr,
  title   = {{TripoSR}: Fast {3D} Object Reconstruction from a Single Image},
  author  = {Tochilkin, Dmitry and Pankratz, David and Guo, Zixuan and Huang, Zifan and Letts, Adam and Li, Yangguang and Liang, Ding and Loy, Chen Change and Varol, Aydin},
  journal = {arXiv preprint arXiv:2403.02151},
  year    = {2024}
}

@inproceedings{xu2024instantmesh,
  title     = {{InstantMesh}: Efficient {3D} Mesh Generation with a Single Image},
  author    = {Xu, Jiale and Hu, Weihao and Geng, Yikun and Liu, Chao and Fang, Jiawei and Zhang, Zhiyuan and Wang, Ziming and Liu, Ying and Shan, Yandong},
  booktitle = {IEEE/CVF Conference on Computer Vision and Pattern Recognition},
  year      = {2024}
}

@article{zhang2024clay,
  title   = {{CLAY}: A Controllable Large-scale Generative Model for Creating High-quality {3D} Assets},
  author  = {Zhang, Longwen and Wang, Ziyu and Zhang, Qixuan and Qiu, Qiwei and Pang, Anqi and Jiang, Haoran and Yang, Wei and Xu, Lan and Yu, Jingyi},
  journal = {ACM Transactions on Graphics},
  volume  = {43},
  number  = {4},
  year    = {2024}
}

@inproceedings{zhao2023michelangelo,
  title     = {{Michelangelo}: Conditional {3D} Shape Generation based on Shape-Image-Text Aligned Latent Representation},
  author    = {Zhao, Zibo and Liu, Wen and Chen, Xin and Zeng, Xianfang and Wang, Rui and Cheng, Pei and Fu, Bin and Chen, Tao and Yu, Gang and Gao, Shenghua},
  booktitle = {Advances in Neural Information Processing Systems},
  year      = {2023}
}

@article{li2025triposg,
  title   = {{TripoSG}: High-Fidelity {3D} Shape Synthesis using Large-Scale Rectified Flow Models},
  author  = {Li, Yangguang and Zou, Zi-Xin and Liu, Zexiang and Wang, Dehu and Liang, Yuan and Yu, Zhipeng and Liu, Xingchao and Guo, Yuan-Chen and Liang, Ding and Ouyang, Wanli and Cao, Yan-Pei},
  journal = {arXiv preprint arXiv:2502.06608},
  year    = {2025}
}

@inproceedings{li2025craftsman3d,
  title     = {{CraftsMan3D}: High-fidelity Mesh Generation with {3D} Native Diffusion and Interactive Geometry Refiner},
  author    = {Li, Weiyu and Liu, Jiarui and Chen, Rui and Liang, Yixun and Chen, Xuelin and Tan, Ping and Long, Xiaoxiao},
  booktitle = {IEEE/CVF Conference on Computer Vision and Pattern Recognition},
  year      = {2025}
}

@article{hunyuan3d2025,
  title   = {{Hunyuan3D} 2.1: From Images to High-Fidelity {3D} Assets with Production-Ready {PBR} Material},
  author  = {{Team Hunyuan3D} and Yang, Shuhui and Yang, Mingxin and Feng, Yifei and Huang, Xin and Zhang, Sheng and He, Zebin and Luo, Di and Liu, Haolin and Zhao, Yunfei and others},
  journal = {arXiv preprint arXiv:2506.15442},
  year    = {2025}
}

@article{li2025step1x3d,
  title   = {{Step1X-3D}: Towards High-Fidelity and Controllable Generation of Textured {3D} Assets},
  author  = {Li, Weiyu and Zhang, Xuanyang and Sun, Zheng and Qi, Di and Li, Hao and Cheng, Wei and Cai, Weiwei and Wu, Shihao and Liu, Jiarui and Wang, Zihao and others},
  journal = {arXiv preprint arXiv:2505.07747},
  year    = {2025}
}

@article{ye2025hi3dgen,
  title   = {{Hi3DGen}: High-Fidelity {3D} Geometry Generation from Images via Normal Bridging},
  author  = {Ye, Chongjie and Wu, Yushuang and Lu, Ziteng and Chang, Jiahao and Guo, Xiaoyang and Zhou, Jiaqing and Zhao, Hao and Han, Xiaoguang},
  journal = {arXiv preprint arXiv:2503.22236},
  year    = {2025}
}

@inproceedings{wu2024direct3d,
  title     = {{Direct3D}: Scalable Image-to-{3D} Generation via {3D} Latent Diffusion Transformer},
  author    = {Wu, Shuang and Lin, Youtian and Zhang, Feihu and Zeng, Yifei and Xu, Jingxi and Torr, Philip and Cao, Xun and Yao, Yao},
  booktitle = {Advances in Neural Information Processing Systems},
  year      = {2024}
}

@article{wu2025direct3ds2,
  title   = {{Direct3D-S2}: Gigascale {3D} Generation Made Easy with Spatial Sparse Attention},
  author  = {Wu, Shuang and Lin, Youtian and Zhang, Feihu and Zeng, Yifei and Yang, Yikang and Bao, Yajie and Qian, Jiachen and Zhu, Siyu and Cao, Xun and Torr, Philip and Yao, Yao},
  journal = {arXiv preprint arXiv:2505.17412},
  year    = {2025}
}

@inproceedings{li2026pixal3d,
  title     = {{Pixal3D}: Pixel-Aligned {3D} Generation from Images},
  author    = {Li, Dong-Yang and Zhao, Wang and Chen, Yuxin and Hu, Wenbo and Guo, Meng-Hao and Zhang, Fang-Lue and Shan, Ying and Hu, Shi-Min},
  booktitle = {ACM SIGGRAPH Conference Papers},
  year      = {2026}
}

@inproceedings{lipman2023flowmatching,
  title     = {Flow Matching for Generative Modeling},
  author    = {Lipman, Yaron and Chen, Ricky T. Q. and Ben-Hamu, Heli and Nickel, Maximilian and Le, Matthew},
  booktitle = {International Conference on Learning Representations},
  year      = {2023}
}

@inproceedings{liu2023rectifiedflow,
  title     = {Flow Straight and Fast: Learning to Generate and Transfer Data with Rectified Flow},
  author    = {Liu, Xingchao and Gong, Chengyue and Liu, Qiang},
  booktitle = {International Conference on Learning Representations},
  year      = {2023}
}

@inproceedings{lan2025gaussiananything,
  title     = {{GaussianAnything}: Interactive Point Cloud Flow Matching for {3D} Object Generation},
  author    = {Lan, Yushi and Zhou, Shangchen and Lyu, Zhaoyang and Hong, Fangzhou and Yang, Shuai and Dai, Bo and Pan, Xingang and Loy, Chen Change},
  booktitle = {International Conference on Learning Representations},
  year      = {2025}
}

@article{chang2024shapetokenization,
  title   = {{3D} Shape Tokenization via Latent Flow Matching},
  author  = {Chang, Jen-Hao Rick and Wang, Yuyang and Bautista Martin, Miguel Angel and Gu, Jiatao and Zhao, Xiaoming and Susskind, Josh and Tuzel, Oncel},
  journal = {arXiv preprint arXiv:2412.15618},
  year    = {2024}
}

@article{zhao2026lato,
  title   = {{LATO}: {3D} Mesh Flow Matching with Structured TOpology Preserving LAtents},
  author  = {Zhao, Tianhao and Zhang, Youjia and Long, Hang and Zhang, Jinshen and Li, Wenbing and Yang, Yang and Zhang, Gongbo and Hladk{\'y}, Jozef and Nie{\ss}ner, Matthias and Yang, Wei},
  journal = {arXiv preprint arXiv:2603.06357},
  year    = {2026}
}

@article{zhang2023shape2vecset,
  title   = {{3DShape2VecSet}: A {3D} Shape Representation for Neural Fields and Generative Diffusion Models},
  author  = {Zhang, Biao and Tang, Jiapeng and Nie{\ss}ner, Matthias and Wonka, Peter},
  journal = {ACM Transactions on Graphics},
  volume  = {42},
  number  = {4},
  year    = {2023}
}

@inproceedings{vahdat2022lion,
  title     = {{LION}: Latent Point Diffusion Models for {3D} Shape Generation},
  author    = {Vahdat, Arash and Williams, Francis and Gojcic, Zan and Litany, Or and Fidler, Sanja and Kreis, Karsten},
  booktitle = {Advances in Neural Information Processing Systems},
  year      = {2022}
}

@inproceedings{lan2024ln3diff,
  title     = {{LN3Diff}: Scalable Latent Neural Fields Diffusion for Speedy {3D} Generation},
  author    = {Lan, Yushi and Hong, Fangzhou and Yang, Shuai and Zhou, Shangchen and Meng, Xuyi and Dai, Bo and Pan, Xingang and Loy, Chen Change},
  booktitle = {European Conference on Computer Vision},
  year      = {2024}
}

@inproceedings{chen2025dora,
  title     = {Dora: Sampling and Benchmarking for {3D} Shape Variational Auto-Encoders},
  author    = {Chen, Rui and Zhang, Jianfeng and Liang, Yixun and Luo, Guan and Li, Weiyu and Liu, Jiarui and Li, Xiu and Long, Xiaoxiao and Feng, Jiashi and Tan, Ping},
  booktitle = {IEEE/CVF Conference on Computer Vision and Pattern Recognition},
  year      = {2025}
}

@article{oquab2023dinov2,
  title={Dinov2: Learning robust visual features without supervision},
  author={Oquab, Maxime and Darcet, Timoth{\'e}e and Moutakanni, Th{\'e}o and Vo, Huy and Szafraniec, Marc and Khalidov, Vasil and Fernandez, Pierre and Haziza, Daniel and Massa, Francisco and El-Nouby, Alaaeldin and others},
  journal={arXiv preprint arXiv:2304.07193},
  year={2023}
}

@article{he2025lam,
  title={Lam: Large avatar model for one-shot animatable gaussian head},
  author={He, Yisheng and Gu, Xiaodong and Ye, Xiaodan and Xu, Chao and Zhao, Zhengyi and Dong, Yuan and Yuan, Weihao and Dong, Zilong and Bo, Liefeng},
  journal={arXiv preprint arXiv:2502.17796},
  year={2025}
}

@inproceedings{ren2024xcube,
  title     = {{XCube}: Large-Scale {3D} Generative Modeling using Sparse Voxel Hierarchies},
  author    = {Ren, Xuanchi and Huang, Jiahui and Zeng, Xiaohui and Museth, Ken and Fidler, Sanja and Williams, Francis},
  booktitle = {IEEE/CVF Conference on Computer Vision and Pattern Recognition},
  year      = {2024}
}

@article{xiong2025octfusion,
  title   = {{OctFusion}: Octree-based Diffusion Models for {3D} Shape Generation},
  author  = {Xiong, Bojun and Wei, Si-Tong and Zheng, Xin-Yang and Cao, Yan-Pei and Lian, Zhouhui and Wang, Peng-Shuai},
  journal = {Computer Graphics Forum},
  volume  = {44},
  number  = {5},
  year    = {2025}
}

@article{chen20243dtopiaxl,
  title   = {{3DTopia-XL}: Scaling High-Quality {3D} Asset Generation via Primitive Diffusion},
  author  = {Chen, Zhaoxi and Tang, Jiaxiang and Dong, Yuhao and Cao, Ziang and Hong, Fangzhou and Lan, Yushi and Wang, Tengfei and Xie, Haozhe and Wu, Tong and Saito, Shunsuke and others},
  journal = {arXiv preprint arXiv:2409.12957},
  year    = {2024}
}

@inproceedings{li2025sparc3d,
  title     = {{Sparc3D}: Sparse Representation and Construction for High-Resolution {3D} Shapes Modeling},
  author    = {Li, Zhihao and Wang, Yufei and Zheng, Heliang and Luo, Yihao and Wen, Bihan},
  booktitle = {Advances in Neural Information Processing Systems},
  year      = {2025}
}

@article{chen2025ultra3d,
  title   = {{Ultra3D}: Efficient and High-Fidelity {3D} Generation with Part Attention},
  author  = {Chen, Yiwen and Li, Zhihao and Wang, Yikai and Zhang, Hu and Li, Qin and Zhang, Chi and Lin, Guosheng},
  journal = {arXiv preprint arXiv:2507.17745},
  year    = {2025}
}

@inproceedings{xiang2025trellis,
  title     = {Structured {3D} Latents for Scalable and Versatile {3D} Generation},
  author    = {Xiang, Jianfeng and Lv, Zelong and Xu, Sicheng and Deng, Yu and Wang, Ruicheng and Zhang, Bowen and Chen, Dong and Tong, Xin and Yang, Jiaolong},
  booktitle = {IEEE/CVF Conference on Computer Vision and Pattern Recognition},
  year      = {2025}
}

@inproceedings{he2025sparseflex,
  title={Sparseflex: High-resolution and arbitrary-topology 3d shape modeling},
  author={He, Xianglong and Zou, Zi-Xin and Chen, Chia-Hao and Guo, Yuan-Chen and Liang, Ding and Yuan, Chun and Ouyang, Wanli and Cao, Yan-Pei and Li, Yangguang},
  booktitle={2025 IEEE/CVF International Conference on Computer Vision (ICCV)},
  pages={14822--14833},
  year={2025},
  organization={IEEE}
}

@article{xiang2026native,
  title   = {Native and Compact Structured Latents for {3D} Generation},
  author  = {Xiang, Jianfeng and Chen, Xiaoxue and Xu, Sicheng and Wang, Ruicheng and Lv, Zelong and Deng, Yu and Zhu, Hongyuan and Dong, Yue and Zhao, Hao and Yuan, Nicholas Jing and Yang, Jiaolong},
  journal = {arXiv preprint arXiv:2512.14692},
  year    = {2025}
}

@article{liu2023one,
  title={One-2-3-45: Any single image to 3d mesh in 45 seconds without per-shape optimization},
  author={Liu, Minghua and Xu, Chao and Jin, Haian and Chen, Linghao and Varma T, Mukund and Xu, Zexiang and Su, Hao},
  journal={Advances in Neural Information Processing Systems},
  volume={36},
  pages={22226--22246},
  year={2023}
}

@inproceedings{long2024wonder3d,
  title={Wonder3d: Single image to 3d using cross-domain diffusion},
  author={Long, Xiaoxiao and Guo, Yuan-Chen and Lin, Cheng and Liu, Yuan and Dou, Zhiyang and Liu, Lingjie and Ma, Yuexin and Zhang, Song-Hai and Habermann, Marc and Theobalt, Christian and others},
  booktitle={2024 IEEE/CVF Conference on Computer Vision and Pattern Recognition (CVPR)},
  pages={9970--9980},
  year={2024},
  organization={IEEE}
}

@inproceedings{riegler2017octnet,
  title={Octnet: Learning deep 3d representations at high resolutions},
  author={Riegler, Gernot and Osman Ulusoy, Ali and Geiger, Andreas},
  booktitle={Proceedings of the IEEE conference on computer vision and pattern recognition},
  pages={3577--3586},
  year={2017}
}

@inproceedings{graham20183d,
  title={3d semantic segmentation with submanifold sparse convolutional networks},
  author={Graham, Benjamin and Engelcke, Martin and Van Der Maaten, Laurens},
  booktitle={2018 IEEE/CVF Conference on Computer Vision and Pattern Recognition},
  pages={9224--9232},
  year={2018},
  organization={Ieee}
}

@inproceedings{choy20194d,
  title={4d spatio-temporal convnets: Minkowski convolutional neural networks},
  author={Choy, Christopher and Gwak, JunYoung and Savarese, Silvio},
  booktitle={2019 IEEE/CVF conference on computer vision and pattern recognition (CVPR)},
  pages={3070--3079},
  year={2019},
  organization={IEEE}
}

@article{yang2021focal,
  title={Focal attention for long-range interactions in vision transformers},
  author={Yang, Jianwei and Li, Chunyuan and Zhang, Pengchuan and Dai, Xiyang and Xiao, Bin and Yuan, Lu and Gao, Jianfeng},
  journal={Advances in neural information processing systems},
  volume={34},
  pages={30008--30022},
  year={2021}
}

@inproceedings{tatarchenko2019single,
  title={What do single-view 3d reconstruction networks learn?},
  author={Tatarchenko, Maxim and Richter, Stephan R and Ranftl, Ren{\'e} and Li, Zhuwen and Koltun, Vladlen and Brox, Thomas},
  booktitle={2019 IEEE/CVF conference on computer vision and pattern recognition (CVPR)},
  pages={3400--3409},
  year={2019},
  organization={IEEE}
}

@inproceedings{wu2018learning,
  title={Learning shape priors for single-view 3d completion and reconstruction},
  author={Wu, Jiajun and Zhang, Chengkai and Zhang, Xiuming and Zhang, Zhoutong and Freeman, William T and Tenenbaum, Joshua B},
  booktitle={European Conference on Computer Vision},
  pages={673--691},
  year={2018},
  organization={Springer}
}

@article{amir2021deep,
  title={Deep vit features as dense visual descriptors},
  author={Amir, Shir and Gandelsman, Yossi and Bagon, Shai and Dekel, Tali},
  journal={arXiv preprint arXiv:2112.05814},
  volume={2},
  number={3},
  pages={4},
  year={2021}
}

@inproceedings{deitke2023objaverse,
  title={Objaverse: A universe of annotated 3d objects},
  author={Deitke, Matt and Schwenk, Dustin and Salvador, Jordi and Weihs, Luca and Michel, Oscar and VanderBilt, Eli and Schmidt, Ludwig and Ehsani, Kiana and Kembhavi, Aniruddha and Farhadi, Ali},
  booktitle={Proceedings of the IEEE/CVF conference on computer vision and pattern recognition},
  pages={13142--13153},
  year={2023}
}

@article{deitke2023objaversexl,
  title={Objaverse-xl: A universe of 10m+ 3d objects},
  author={Deitke, Matt and Liu, Ruoshi and Wallingford, Matthew and Ngo, Huong and Michel, Oscar and Kusupati, Aditya and Fan, Alan and Laforte, Christian and Voleti, Vikram and Gadre, Samir Yitzhak and others},
  journal={Advances in Neural Information Processing Systems},
  volume={36},
  pages={35799--35813},
  year={2023}
}

@inproceedings{lai2026lattice,
  title     = {{LATTICE}: Democratize High-Fidelity {3D} Generation at Scale},
  author    = {Lai, Zeqiang and Zhao, Yunfei and Zhao, Zibo and Liu, Haolin and Lin, Qingxiang and Huang, Jingwei and Guo, Chunchao and Yue, Xiangyu},
  booktitle = {IEEE/CVF Conference on Computer Vision and Pattern Recognition},
  year      = {2026}
}

@inproceedings{liu2021swin,
  title     = {{Swin Transformer}: Hierarchical Vision Transformer using Shifted Windows},
  author    = {Liu, Ze and Lin, Yutong and Cao, Yue and Hu, Han and Wei, Yixuan and Zhang, Zheng and Lin, Stephen and Guo, Baining},
  booktitle = {IEEE/CVF International Conference on Computer Vision},
  year      = {2021}
}

@inproceedings{xue2024ulip,
  title={Ulip-2: Towards scalable multimodal pre-training for 3d understanding},
  author={Xue, Le and Yu, Ning and Zhang, Shu and Panagopoulou, Artemis and Li, Junnan and Mart{\'\i}n-Mart{\'\i}n, Roberto and Wu, Jiajun and Xiong, Caiming and Xu, Ran and Niebles, Juan Carlos and others},
  booktitle={2024 IEEE/CVF Conference on Computer Vision and Pattern Recognition (CVPR)},
  pages={27081--27091},
  year={2024},
  organization={IEEE}
}

@inproceedings{zhou2024uni3d,
  title={Uni3d: Exploring unified 3d representation at scale},
  author={Zhou, Junsheng and Wang, Jinsheng and Ma, Baorui and Liu, Yu-Shen and Huang, Tiejun and Wang, Xinlong},
  booktitle={International Conference on Learning Representations},
  year={2024}
}

@inproceedings{stojanov2021using,
  title     = {Using Shape to Categorize: Low-Shot Learning with an Explicit Shape Bias},
  author    = {Stojanov, Stefan and Thai, Anh and Rehg, James M.},
  booktitle = {IEEE/CVF Conference on Computer Vision and Pattern Recognition},
  year      = {2021}
}

@misc{tripov31,
  author       = {{Tripo AI}},
  title        = {Tripo Studio: Tripo v3.1},
  year         = {2026},
  howpublished = {\url{https://studio.tripo3d.ai/}},
  note         = {Accessed: September 23, 2026}
}

@misc{hunyuan3d31,
  author       = {{Tencent Hunyuan}},
  title        = {Hunyuan3D 3.1},
  year         = {2026},
  howpublished = {\url{https://3d.hunyuan.tencent.com/}},
  note         = {Accessed: September 23, 2026}
}

@misc{rodinv25,
  author       = {{Hyper3D}},
  title        = {Rodin v2.5},
  year         = {2026},
  howpublished = {\url{https://hyper3d.ai/workspace/rodin}},
  note         = {Accessed: September 23, 2026}
}
\bibliographystyle{iclr2027_conference}

\appendix
\section{Appendix}
\subsection{Model Details}

\paragraph{Configurations inherited from TRELLIS.2.}
\textsc{Filigree3D} builds on the SLat flow-matching architecture of TRELLIS.2~\citep{xiang2026native} and retains selected architectural and training configurations. The denoiser processes 32-channel structured latent features using 30 Transformer blocks with an MLP expansion ratio of \(5.3334\). Self-attention uses rotary positional embeddings (RoPE). We optimize the model using AdamW with a learning rate of \(1\times10^{-4}\).
For reproducibility, Table~\ref{tab:network-details} summarizes the architectural and training configurations of our sparse latent flow-matching model.
\paragraph{Filigree3D-specific configurations.}
The denoiser uses a hidden width of 2,048 and 16 attention heads, resulting in 2.2B parameters. For Structure-Aware Sparse Scaling, we set the token budget to \(B=50{,}000\). Occupancy-Balanced Core--Halo Cropping uses a coarse-cell width of \(g=8\) and a halo width of \(h=2\). For multi-level conditioning, we concatenate DINOv2 features from layers 5, 7, 11, and 23 along the channel dimension~\citep{he2025lam}. The resulting representation conditions every DiT block via cross-attention. Local attention uses a window width of \(w=32\), while coarse-global communication is applied at an interval of \(K=6\) DiT blocks with a pooling stride of \(p=8\).

\subsection{Implementation Details}
\paragraph{SC-VAE.} To support reliable geometry encoding and decoding at high resolutions, we fine-tune the SC-VAE separately for each target resolution and address two resolution-dependent limitations in TRELLIS.2. 

(1) Flexible-Dual-Grid represents flattened voxel indices using 32-bit keys for neighbor lookup. A $2048^3$ grid contains $2^{33}$ possible voxel positions, exceeding the 32-bit key space. The resulting key collisions may incorrectly connect distant voxels and produce spurious long edges. We therefore use 64-bit keys whenever $R^3>2^{32}$, where $R$ denotes the grid resolution. 

(2) The \texttt{flex\_gemm} backend uses 10 bits per coordinate, restricting each axis to $[0,1023]$. To process a $4096^3$ grid, we divide it into $4\times4\times4$ non-overlapping blocks of size $1024^3$ and encode each block independently using local coordinates. With $16\times$ spatial downsampling, each block yields a $64^3$ latent block. We then assemble these blocks into a $256^3$ latent grid by adding the offset $64\mathbf{b}$, where $\mathbf{b}\in\{0,1,2,3\}^3$ denotes the block index. This strategy applies more generally when the block size does not exceed 1024 and is divisible by 16.

\paragraph{DiT.}
We train the denoiser on 16 NVIDIA H20 GPUs with a global batch size of 16 and a learning rate of $1\times10^{-4}$. We first train it on TRELLIS-500K, then fine-tune the $2048^3$ model on the high-detail subset selected by our resolution-gain-aware data curation strategy, together with an additional collection of 200K objects. We additionally fine-tune the sparse-structure DiT to generate structures on a $128^3$ voxel grid. At inference, we use 12 sampling steps.

\begin{table}[t]
    \centering
    \small
    \setlength{\tabcolsep}{5pt}
    \caption{Configuration of the sparse latent flow-matching denoiser.}
    \label{tab:network-details}
    \begin{tabular}{@{}llll@{}}
        \toprule
        \textbf{Setting} & \textbf{Value}
        & \textbf{Setting} & \textbf{Value} \\
        \midrule
        Input image resolution
        & \(1024\times1024\)
        & Local window width \(w\)
        & 32 \\

        Image encoder
        & DINOv2 ViT-L/16
        & Global communication interval \(K\)
        & 6 blocks \\

        Feature dimension per layer
        & 1,024
        & Coarse pooling stride \(p\)
        & 8 \\

        DINOv2 feature layers
        & 5, 7, 11, 23
        & Positional encoding
        & RoPE \\

        SLat channels
        & 32
        & Attention backend
        & FlashAttention-3 \\

        SLat / geometry resolution
        & \(128^3\) / \(2048^3\)
        & Optimizer
        & AdamW \\

        Token budget \(B\)
        & 50,000
        & Learning rate
        & \(1\times10^{-4}\) \\

        Coarse-cell width \(g\) / halo width \(h\)
        & 8 / 2
        & Batch size per GPU
        & 1 \\

        Transformer width / depth
        & 2,048 / 30
        & Training steps
        & 1,000,000 \\

        Attention heads
        & 16
        & Precision
        & bfloat16 \\

        MLP expansion ratio
        & 5.3334
        & EMA decay
        & 0.9999 \\

        Denoiser parameters
        & 2.2B
        & CFG dropout probability
        & 0.1 \\
        \bottomrule
    \end{tabular}
\end{table}
\subsubsection{Geometry Metrics}
\label{app:geo}
We evaluate \textsc{Filigree3D} using four metrics: Chamfer Distance (CD), F-scores at thresholds of \(0.01\) and \(0.002\), ULIP-2 similarity~\citep{xue2024ulip}, and Uni3D similarity~\citep{zhou2024uni3d}. The specific computation of each metric is defined as follows.
\paragraph{Visible-surface extraction.}
Generated meshes may contain floating components or internal shells that are not externally observable. We therefore evaluate only the outer surface. For a mesh $\mathcal{M}$ with face set $\mathcal{F}$, we uniformly sample directions on an enclosing sphere and cast $R=20{,}000$ rays toward the mesh using the Open3D \texttt{RaycastingScene}. A face is considered visible if it is intersected by at least one ray. The union of these faces defines the visible submesh $\mathcal{M}_{\mathrm{vis}}$. We compute its relative area as
\begin{equation}
\frac{\mathrm{Area}(\mathcal{M}_{\mathrm{vis}})}
{\mathrm{Area}(\mathcal{M})}.
\end{equation}
\paragraph{Surface sampling.}
We independently sample $N=100{,}000$ points from the visible submeshes of the prediction and ground truth using area-weighted uniform sampling. This yields the predicted and reference point clouds $P$ and $G$, respectively.
\paragraph{Chamfer distance.}
For a point $\mathbf{x}$ and point set $S$, we define the nearest neighbor distance as
\begin{equation}
d(\mathbf{x},S)=\min\nolimits_{\mathbf{y}\in S}\|\mathbf{x}-\mathbf{y}\|_2,
\end{equation}
which is computed using a KD-tree. We evaluate the untrimmed bidirectional Chamfer distances as
\begin{align}
\mathrm{CD}_{\ell_1}
&=
\frac{1}{|G|}
\sum\nolimits_{\mathbf{p}\in G}d(\mathbf{p},P)
+
\frac{1}{|P|}
\sum\nolimits_{\mathbf{q}\in P}d(\mathbf{q},G),\\
\mathrm{CD}_{\ell_2}
&=
\frac{1}{|G|}
\sum\nolimits_{\mathbf{p}\in G}d(\mathbf{p},P)^2
+
\frac{1}{|P|}
\sum\nolimits_{\mathbf{q}\in P}d(\mathbf{q},G)^2.
\end{align}
\begin{figure}[t]
    \centering
    \includegraphics[width=\linewidth]{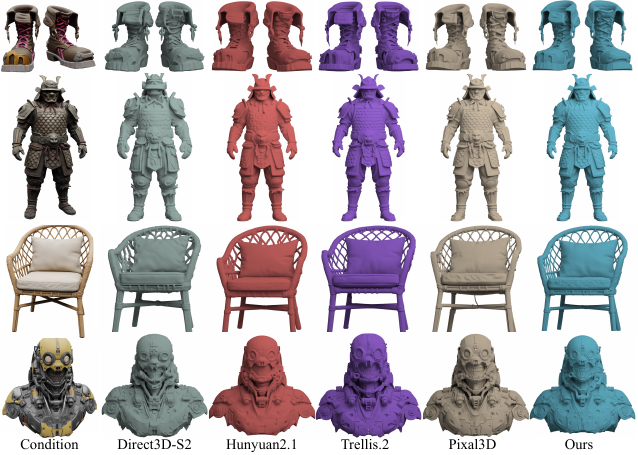}
    \caption{\textbf{More qualitative comparisons on FineGeo3D dataset.} Our method reconstructs finer geometric structures and preserves intricate details more faithfully than existing methods.}
    \label{fig:more_vis_results}
\end{figure}

We also report the two directional mean distances:
\begin{align}
\bar d_{G\rightarrow P}
&=
\frac{1}{|G|}
\sum\nolimits_{\mathbf{p}\in G}d(\mathbf{p},P),\\
\bar d_{P\rightarrow G}
&=
\frac{1}{|P|}
\sum\nolimits_{\mathbf{q}\in P}d(\mathbf{q},G).
\end{align}

\paragraph{F-score.}
Given a distance threshold $\tau$, we define precision and recall as
\begin{align}
\mathrm{Prec}_{\tau}
&=
\frac{1}{|P|}
\sum\nolimits_{\mathbf{q}\in P}
\mathbf{1}\!\left[d(\mathbf{q},G)<\tau\right],\\
\mathrm{Rec}_{\tau}
&=
\frac{1}{|G|}
\sum\nolimits_{\mathbf{p}\in G}
\mathbf{1}\!\left[d(\mathbf{p},P)<\tau\right].
\end{align}
Their harmonic mean is
\begin{equation}
F_{\tau}
=
\frac{2\,\mathrm{Prec}_{\tau}\mathrm{Rec}_{\tau}}
{\mathrm{Prec}_{\tau}+\mathrm{Rec}_{\tau}}.
\end{equation}
We set $F_{\tau}=0$ when both precision and recall are zero. We report $F_{0.01}$ and the stricter $F_{0.002}$. After similarity alignment, all objects share the ground-truth scale, making these absolute thresholds comparable across methods.

\paragraph{ULIP-2 and Uni3D scores.}
We encode each generated point cloud using either the ULIP-2 PointBERT encoder or the Uni3D point encoder. The corresponding CLIP vision encoder embeds the reference image, which is also used to condition all methods. For each model, we compute the semantic score as the cosine similarity between the normalized point-cloud and image embeddings:
\begin{equation}
\hat{\mathbf{z}}_{u}
=
\frac{\mathbf{z}_{u}}{\|\mathbf{z}_{u}\|_2},
\quad
u\in\{\mathrm{pc},\mathrm{img}\},
\qquad
S_{\mathrm{sem}}
=
\hat{\mathbf{z}}_{\mathrm{pc}}^{\top}
\hat{\mathbf{z}}_{\mathrm{img}}.
\end{equation}
\subsection{Additional Comparisons and $2048^3$ Generation Results}
\begin{figure}[!t]
    \includegraphics[width=\linewidth]{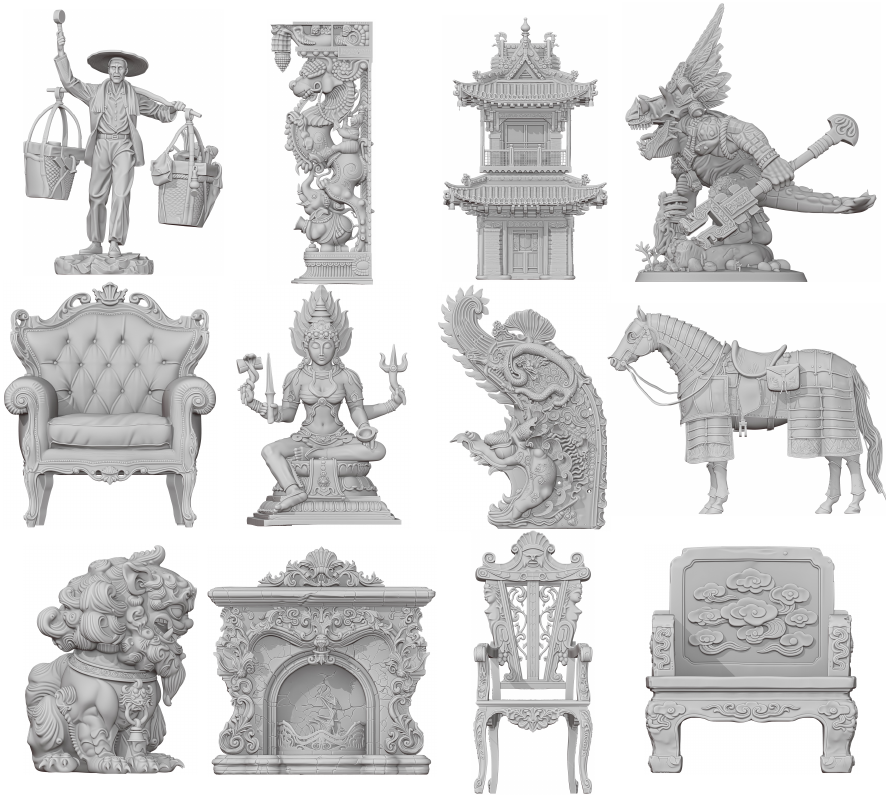}
    \caption{\textbf{Additional $2048^3$ generation results.} Our FineGeo3D recovers fine-grained geometric details while preserving coherent global structures across diverse objects.}
    \label{fig:more_2048}
\end{figure}

We provide additional qualitative comparisons with Direct3D-S2~\citep{wu2024direct3d}, Hunyuan3D 2.1~\citep{hunyuan3d2025}, TRELLIS.2~\citep{xiang2026native}, and Pixal3D~\citep{li2026pixal3d} on the FineGeo3D dataset. As shown in Figure~\ref{fig:more_vis_results}, our method better recovers fine local geometry and thin structures while maintaining coherent global shapes across diverse object categories. These comparisons further illustrate its ability to represent geometrically complex regions that are easily smoothed or omitted by existing methods. Figure~\ref{fig:more_2048} presents additional $2048^3$ generation results, including objects with intricate surfaces and slender components. The results show that increasing the generation resolution enables the recovery of finer geometric structures without compromising overall shape consistency.

\end{document}